\documentclass[11pt]{article}

\usepackage[final]{acl}

\usepackage{times}
\usepackage{latexsym}
\usepackage{pifont}
\usepackage{amsmath}
\usepackage{amssymb}
\usepackage{amsfonts}
\usepackage{array}
\usepackage{hyperref}

\usepackage[T1]{fontenc}

\usepackage[utf8]{inputenc}

\usepackage{microtype}

\usepackage{inconsolata}

\usepackage{graphicx}
\usepackage{subcaption}
\usepackage{multirow}
\usepackage{array}
\usepackage{booktabs}
\usepackage{colortbl}
\usepackage{tcolorbox}
\usepackage{listings}
\usepackage{algorithm}
\usepackage{algpseudocode} 
\usepackage{enumitem}
\usepackage{amsmath}
\usepackage{tabularx}

\usepackage{seqsplit}

\newcolumntype{Y}{>{\raggedright\arraybackslash}X}

\tcbuselibrary{breakable,skins}

\definecolor{speco3blue}{RGB}{222,234,246}

\title{Spec-o3: A Tool-Augmented Vision-Language Agent for Rare Celestial Object Candidate Vetting via Automated Spectral Inspection}

\author{
 \textbf{Minghui Jia}\textsuperscript{\normalfont 1,2}\quad
 \textbf{Qichao Zhang}\textsuperscript{\textnormal{1,4\textdagger}}\quad
 \textbf{Ali Luo}\textsuperscript{\normalfont {3\textdagger}}\quad
 \textbf{Linjing Li}\textsuperscript{\normalfont {1,4}}\quad
 \\
 \textbf{Shuo Ye}\textsuperscript{\normalfont 3}\quad
 \textbf{Hailing Lu}\textsuperscript{\normalfont 3}\quad
 \textbf{Wen Hou}\textsuperscript{\normalfont 3}\quad
 \textbf{Dongbin Zhao}\textsuperscript{\normalfont 1,2,4}
\\
 \textsuperscript{1}SKL-MAIS, Institute of Automation, CAS, Beijing, China \\
 \textsuperscript{2}School of Advanced Interdisciplinary Sciences, UCAS, Beijing, China \\
 \textsuperscript{3}National Astronomical Observatories, CAS, Beijing, China \\
 \textsuperscript{4}School of Artificial Intelligence, UCAS, Beijing, China \\
 {\ttfamily \{jiaminghui2025,zhangqichao2014\}@ia.ac.cn}
}

\begin{document}
\maketitle
\begin{abstract}

Due to the limited generalization and interpretability of deep learning classifiers, the final vetting of rare celestial object candidates still relies on manually intensive expert visual inspection, which has become a primary bottleneck as modern spectroscopic surveys continue to scale.
To bridge this gap, we propose Spec-o3, a tool-augmented vision-language agent that performs astronomer-aligned spectral inspection via interleaved multimodal chain-of-thought reasoning.
Spec-o3 is trained with a two-stage post-training recipe: cold-start supervised fine-tuning on expert inspection trajectories followed by outcome-based reinforcement learning on rare-type verification tasks.
Evaluated on five rare-object identification tasks from LAMOST, Spec-o3 establishes a new State-of-the-Art, boosting the macro-F1 score from 28.3 to 76.5 with a 7B parameter base model and outperforming both proprietary VLMs and specialized deep models. Beyond accuracy, Spec-o3 processes spectra at $\sim$0.2\,s per sample on an 8$\times$H100 server, a $\sim$50$\times$ throughput gain over expert manual inspection. The agent also demonstrates strong generalization to unseen inspection tasks across survey shifts (from LAMOST to SDSS/DESI). Expert evaluations further confirm that its reasoning traces are coherent and physically consistent, supporting transparent and trustworthy decision-making.
Code, data, and models are available at \href{https://github.com/Maxwell-Jia/spec-o3}{Project HomePage}.

\end{abstract}

\section{Introduction}

\begin{figure}[t]
  \centering
  \includegraphics[width=\linewidth]{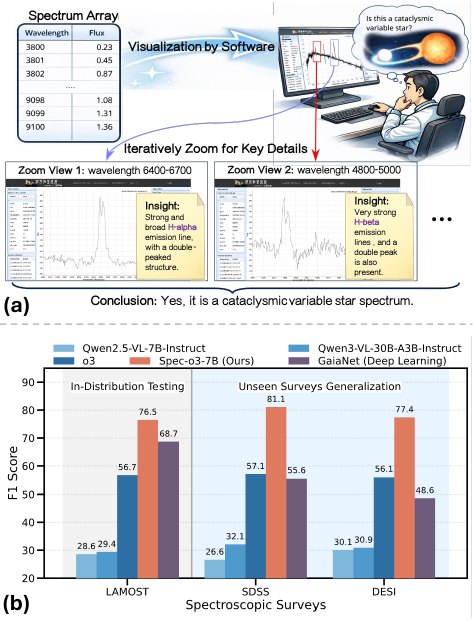}
  \caption{\textbf{(a) Astronomers' visual inspection workflow.} Astronomers visualize the raw numerical arrays to assess global morphology, then iteratively zoom into specific wavelength regions to examine fine-grained features for the final decision. \textbf{(b) Performance comparison.} Spec-o3 achieves state-of-the-art performance and good generalization across all datasets.}
  \label{fig:workflow-performance}
\end{figure}

\begin{figure}[t]
  \centering
  \includegraphics[width=\linewidth]{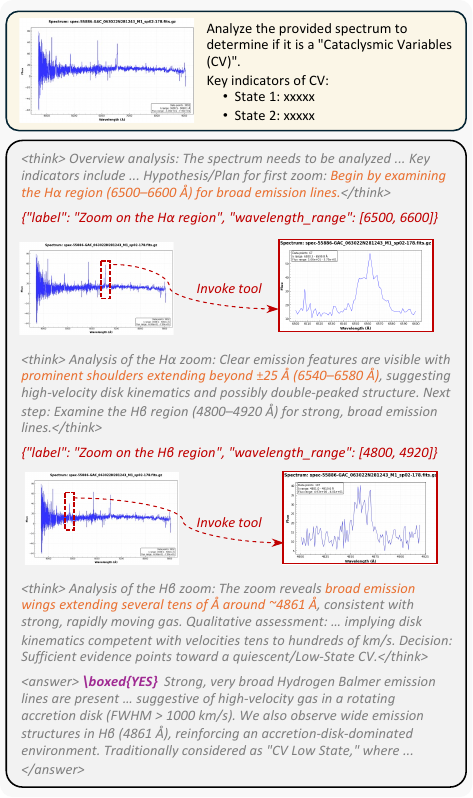}
  \caption{An illustration of Spec-o3's Interleaved Multimodal Chain-of-Thought. The agent iteratively alternates between textual reasoning (<think>…</think>) and fine-grained visual evidence from tool-rendered zoomed spectrum plots. Red JSON shows the tool calls. The final decision is in <answer>...</answer>.}
  \label{fig:case-study}
\end{figure}

Detecting rare celestial objects and establishing catalogs for them is one of the core objectives of many large-scale spectroscopic surveys \citep{york2000sdss,cui2012lamost,aghamousa2016desi}, as these long-tail objects are instrumental in deepening and refining our understanding of astrophysical theories. 
In practice, developing such catalogs generally involves a two-stage process that combines automated candidate screening with expert vetting \citep{tan2025robust,kong2021identification,inight2025cataclysmic}. 
Deep learning algorithms can scale the screening stage over massive spectral archives \citep{lanusse2023dawes}, but the final vetting stage still relies on manual visual inspection to filter subtle contaminants and instrumental artifacts and ensure catalog reliability \citep{kim2024accurate,borra2015validation,lan2023desi}.
However, the vetting stage is difficult to scale.
For instance, building a cataclysmic variable catalog from LAMOST survey required experts to visually inspect about 170k candidates screened from roughly 10 million archived spectra, ultimately confirming only 323 objects \citep{sun2021catalog}. 
With the exponential data growth of next-generation surveys, candidate volumes will continue to surge \citep{fulmer2023astro,vincent2023data,li2023spectrumva}, making manual inspection a major bottleneck in modern astronomy \citep{fluke2020understanding,fluke2023survey}.

A key reason why manual inspection is still required is that deep learning models typically produce opaque probability scores and exhibit limited out-of-distribution generalization, which undermines expert trust \citep{lieu2025comprehensive,wetzel2025interpretable}.
Although post-hoc methods such as Grad-CAM \citep{selvaraju2017grad}, Integrated Gradients \citep{sundararajan2017axiomatic}, LIME \citep{ribeiro2016should}, and SHAP \citep{lundberg2017unified} offer some interpretability, they mainly produce coarse feature attributions. Such noisy explanations often cannot be reliably mapped onto astrophysical structures \citep{stoppa2025textual,lieu2025comprehensive}, rendering them insufficient to substitute for expert inspection.

We observe that real-world inspection is a ``think-with-spectral-image'' process as illustrated in Figure \ref{fig:workflow-performance}(a).
Astronomers typically rely on software to visualize spectra stored as numerical arrays, whereby they form an initial judgment from the global morphology and subsequently iteratively zoom into task-relevant wavelength regions to verify local details. 
Finally, they make a vetting decision on whether to include the candidate in the catalog.
The software details are provided in Appendix \ref{appx:software}.
This raises a key question: \textbf{Can we design an expert-trusted  and highly generalized vetting agent to inspect spectra like Astronomers?}

Aligning models with the iterative expert workflow may improve acceptance of automated vetting.
Vision-Language Models (VLMs) have shown promise in generating expert-level explanations for transient imaging classification \citep{stoppa2025textual}. 
Building on this insight, a natural direction is to augment VLMs with a spectral visualization tool (see Appendix \ref{appx:tool_details} for details), so that they can inspect spectra like Astronomers in the vetting stage.
However, even advanced think-with-image models such as o3 \citep{o3} perform poorly on this task (Figure \ref{fig:workflow-performance}(b)), failing to consistently distinguish subtle spectral-shape differences. While task-specific fine-tuning is a natural remedy, it is often impractical because it requires a large amount of fully annotated expert trajectory data.

To overcome these limitations, we introduce \textbf{Spec-o3}, a tool-augmented agent for rare celestial object candidate vetting via automated spectral inspection. 
Spec-o3 follows an Interleaved Multimodal Chain-of-Thought (iMCoT) trajectory (Figure \ref{fig:case-study}), alternating between textual reasoning and fine-grained visual evidence from tool-rendered zoomed spectral views before producing a final vetting decision.
It adopts a two-stage post-training strategy that cold-starts with a small set of expert trajectories and then scales via outcome-based reinforcement learning \citep{fu2026srft}.

Our main contributions are as follows:
\begin{itemize}
  \item We establish a rigorous, expert-driven data construction pipeline and curate a high-quality astronomy iMCoT dataset to provide a reliable cold-start for automated spectral inspection.
  \item We propose Spec-o3, an end-to-end tool-augmented VLM agent that achieves state-of-the-art performance (76.5\% macro-F1) on rare celestial object vetting, significantly outperforming specialized deep models and proprietary VLMs.
  \item We demonstrate the practical application value of Spec-o3, showcasing robust zero-shot generalization across unseen surveys and inspection tasks, and a $\sim$50$\times$ inference throughput gain over expert manual inspection that directly relieves the vetting bottleneck of modern spectroscopic workflows.
\end{itemize}

\section{Related Works}

\subsection{Cataloging Rare Celestial Objects}
Research on rare celestial object cataloging can be broadly grouped into screening and vetting. 
Machine learning and deep learning are now standard for screening and can retrieve rare-object candidates at archive scale \citep{he2024carbonnet,zhang2025white,inight2025cataclysmic,tan2025robust,fang2025catalog,jia2025fcn4flare,zhu2025flare}.
Yet vetting still relies on expert visual inspection as the final quality-control gate, filtering artifacts and false positives while providing high-fidelity labels for downstream use \citep{lan2023desi,alexander2023desi,rojas2023impact}.
Prior works therefore focus on software and interfaces that streamline manual inspection to help experts surface key evidence and standardize vetting workflows \citep{juneau2024sparcl,landriau2025desi}, such as MARZ \citep{hinton2016marz}, ASERA \citep{yuan2016team}, SpectrumVA \citep{li2023spectrumva} and Prospect \citep{ratajczak2025compilation,juneau2025identifying}.
However, the literature still lacks end-to-end methods that automate expert vetting with an auditable, workflow-aligned inspection process.

\subsection{Multimodal Large Language Models}
\label{sec:mllm}

Recent years have seen rapid advances in multimodal large language models (MLLMs). Early approaches typically paired pretrained vision encoders with LLMs using lightweight adapters or projection modules, enabling basic cross-modal alignment and simple multimodal reasoning \citep{Liu2023LlaVA,Li2023BLIP2BL,bai2023qwen,Liu_2024_CVPR,chen2024internvl,tu2025perception}. As research progressed, models such as Qwen2.5-VL \citep{bai2025qwen25vl}, LLaVA-OneVision \citep{li2024llava} and InternVL3 \citep{zhu2025internvl3} scaled up both data and model capacity, leading to substantial improvements on several tasks. More recent OmniMLLM systems \citep{fu2024vita,hong2025worldsense,li2025baichuanomni15,zhao2025r1omni} further extend the range of supported modalities, jointly processing images, video, speech, and other signals within a unified framework. Nevertheless, these models are still largely used as passive perception engines that map multimodal inputs to textual responses, and seldom engage in domain-specific, tool-augmented analysis of scientific data.

\subsection{Think with Image}
\label{sec:think_with_image}

The think-with-image paradigm, presented first by o3 \citep{o3}, has inspired a series of open-source efforts to reproduce and extend tool-augmented visual reasoning.
GRIT \citep{fan2025grit} and Pixel Reasoner \citep{pixel-reasoner_wang2025} explicitly integrate visual information to achieve a more precise visual focus. VLM-R$^3$ \citep{jiang2025vlm}, Chain-of-Focus \citep{Chain-of-Focus_zhang2025} and Mini-o3 \citep{lai2025minio3} employ two-stage training pipelines to enhance tool-use capabilities. DeepEyes \citep{zheng2025deepeyes} explores the use of reinforcement learning alone to enhance the model's ability to think with images. To improve generality, PyVision \citep{zhao2025pyvision} and Thyme \citep{zhang2025thyme} introduce the execution of programmatic code into the visual reasoning loop, allowing flexible manipulation of visual operations. DeepEyes-v2 \citep{hong2025deepeyesv2} further extends the tool set by searching, enabling models to retrieve external knowledge. 
Unlike generic think-with-image VLMs focused on natural images, Spec-o3 emphasizes scientific knowledge alignment for spectra, linking abstract diagnostics to visual evidence in numerical data.

\begin{figure*}[t]
  \centering
  \includegraphics[width=\textwidth]{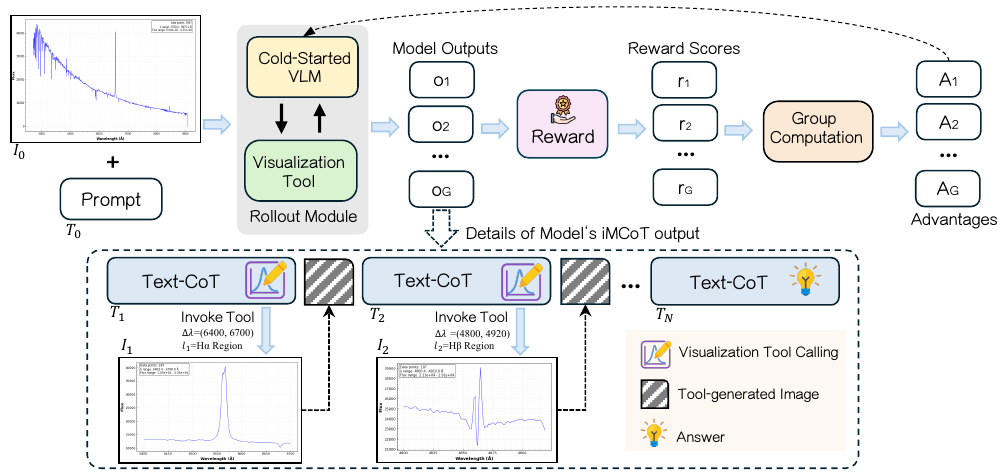}
  \caption{Overview of the Spec-o3 framework. Given a prompt $T_0$ and an initial view $I_0$, the VLM generates an iMCoT trajectory in which text reasoning blocks $T_n$ are interleaved with tool-generated images parameterized by wavelength interval $\Delta\lambda$ and optional label $l_n$, until the final text output $T_N$ is produced. The VLM is cold-start initialized and optimized with GRPO.}
  \label{fig:overview}
\end{figure*}

\section{Method}

\subsection{Spec-o3}
\label{sec:spec-o3}

Spec-o3 is a tool-augmented multimodal agent that performs ``think-with-spectral-image'' inspection.
It is built upon Qwen2.5-VL \citep{bai2025qwen25vl} and trained through a two-stage post-training procedure, combining cold-start (Section \ref{sec:cold_start}) with agentic reinforcement learning (Section \ref{sec:agentic_rl}).

As illustrated in Figure \ref{fig:overview}, Spec-o3 follows an iMCoT trajectory, alternating between textual reasoning and tool-generated spectral images.
Spec-o3 takes as input a text prompt $T_0$ and an initial visualization $I_0$ rendered from the original spectral array.
Here, $T_0$ includes a discriminative query about whether the spectrum matches the target category, along with an expert-written summary of its diagnostic features and common contaminants.

To formalize the iterative reasoning process, we define the state $s_t$ of iMCoT at step t as follows:
\begin{equation}
s_t = \big\{ (I_k, T_k) \big\}_{k=0}^{t}
    = \{ I_{\le t}, T_{\le t} \},
\end{equation}
where $I_{\le t} = \{I_0, I_1, \ldots, I_t\}$ denotes the set of rendered spectral visualizations observed so far, and $T_{\le t} = \{T_0, T_1, \ldots, T_t\}$ represents the corresponding textual tokens.
Given the current state $s_t$, the action $a_t \sim \pi_\theta(a \mid s_t)$ is drawn from the policy $\pi_\theta$.
Specifically, Spec-o3 autonomously determines whether to directly output a final answer or to use the spectral visualization tool $Tool_t$ to acquire more fine-grained evidence from wavelength regions of interest.
Here, $Tool_t$ denotes the tool invocation at step $t$, whose input is a wavelength interval $\Delta\lambda_t = (\lambda_t^{\min}, \lambda_t^{\max})$ accompanied by an optional textual label $l_t$ for the queried diagnostic region.
Upon successful execution, the tool returns a localized re-rendering $I_{t+1}$ restricted to $\Delta\lambda_t$, which is appended to the state as the next observation.
This interaction repeats until the model outputs the final text block $T_N$ containing the answer, or reaches a preset limit on tool invocations.
The resulting iMCoT trajectory can be written as
\begin{equation}
\tau = \big( T_0, I_0, T_1, I_1, T_2, I_2, \ldots, T_N \big).
\end{equation}

\subsection{Cold Start}
\label{sec:cold_start}

To initiate the cold start phase, we sample spectra for the five tasks, including Cataclysmic Variables (CV), Carbon Stars (CS), S-type Stars (SS), M-type Giants (MG), and White Dwarfs (WD), from the corresponding LAMOST official catalogs\footnote{\url{https://www.lamost.org/dr11/v2.0/catalogue}}. We additionally sample spectra from several confusing types selected by astronomers based on domain experience. After filtering all sampled spectra by Signal-to-Noise Ratio (SNR) $>$ 10, we obtain an initial pool of approximately 4k spectra for trajectory construction.

Astronomers first formulate simplified inspection guidelines for each task, specifying key spectral features and typical contamination patterns. 
We then prompt GPT-5 \citep{gpt5} with these guidelines and ground-truth labels, enabling it to invoke spectral visualization tool to generate initial reasoning trajectories. 
Astronomers screen these drafts and discard trajectories that are evidently implausible. 
For each remaining draft, one astronomer performs a focused revision by correcting tool arguments and rewriting the accompanying analysis to ensure that each claim is supported by the rendered evidence. 
Two additional astronomers then independently audit the revised trajectory against the same guidelines. 
If either auditor requests changes, the trajectory is returned for revision and re-audited until all three experts approve it. 
The approved trajectories are standardized by GPT-5 into iMCoT-formatted reasoning traces optimized for learning. 
After rewriting, the three astronomers jointly perform a final acceptance vote on each trajectory to confirm its fidelity to the validated tool actions and evidence statements. 
Trajectories that fail this final vote are discarded. After this process, we compile a cold start dataset of approximately 1k high-quality trajectories.

Subsequently, we performed supervised fine-tuning (SFT) on Qwen2.5-VL \citep{bai2025qwen25vl}. To prevent the model from memorizing tool outputs, we applied a token-level loss mask to the tool returns \citep{fu2026opd}. This strategy encourages the model to learn tool invocation and interpretation methodologies rather than memorizing visualization results.

\subsection{Agentic Reinforcement Learning}
\label{sec:agentic_rl}

The cold start phase injects basic expert priors and enables stable tool use.
However, performance is still constrained by the scarcity of full expert trajectories,
especially in noisy and heavily contaminated cases.
We therefore apply outcome-based reinforcement learning after cold start, using relatively abundant label-only data to further improve performance \citep{tu2025autothink,tu2025dpo}.

\paragraph{Optimization}
As illustrated in Figure \ref{fig:overview}, we utilize Group Relative Policy Optimization (GRPO) \citep{shao2024deepseekmath} for reinforcement learning. Consistent with the cold start phase, to prevent tool outputs from interfering with the training objective, we apply a token-wise loss mask to all response tokens of tool, effectively excluding them from the loss computation.

\paragraph{Reward Design}
In the agentic reinforcement learning phase, we use a rule-based outcome reward to evaluate each trajectory, following \citet{deepseek-r1}. The reward prioritizes final prediction accuracy and enforces output-format constraints. Since tool use is already reliable after cold start, we do not add an explicit tool-usage reward term as in \citet{zheng2025deepeyes}.
Given a reasoning trajectory $\tau$, the reward function is defined as:
\begin{equation}
\label{eq:reward_function}
r(\tau) =
\begin{cases}
1, & \text{if } o_{\text{pred}} = o_{\text{gold}} \;\wedge\; f_{\text{fmt}}(y), \\
1 - \alpha, & \text{if } o_{\text{pred}} = o_{\text{gold}} \;\wedge\; \neg f_{\text{fmt}}(y), \\
0, & \text{if } o_{\text{pred}} \neq o_{\text{gold}} \;\wedge\; f_{\text{fmt}}(y), \\
-\alpha, & \text{if } o_{\text{pred}} \neq o_{\text{gold}} \;\wedge\; \neg f_{\text{fmt}}(y),
\end{cases}
\end{equation}
where $o_{\text{pred}}$ and $o_{\text{gold}}$ are the predicted and ground-truth labels, $f_{\text{fmt}}(y)$ indicates whether the response $y$ satisfies the required formatting constraints, and $\alpha$ controls the penalty for format violations.

\begin{table*}[t]
  \centering
  \caption{\textbf{Main results on SpecVI-Bench} across five tasks: Cataclysmic Variables (CV), Carbon Stars (CS), S-type Stars (SS), M-type Giants (MG), and White Dwarfs (WD). We report Accuracy (Acc) and F1 for each task and their macro-average. All Qwen-VL baselines use the Instruct variants. Spec-o3 is trained from Qwen2.5-VL. Best and second-best results are in \textbf{bold} and \underline{underlined}, respectively.}
  \label{tab:main_results}
  \small
  {
  \renewcommand{\arraystretch}{1.2}

  \begin{tabular*}{\textwidth}{@{\extracolsep{\fill}} %
    l |
    c@{\hspace{5pt}}c
    c@{\hspace{5pt}}c 
    c@{\hspace{5pt}}c 
    c@{\hspace{5pt}}c 
    c@{\hspace{5pt}}c |
    c@{\hspace{5pt}}c  
  }
  \toprule
  \multirow{2}{*}{\textbf{Model}} &
  \multicolumn{2}{c}{\textbf{CV}} &
  \multicolumn{2}{c}{\textbf{CS}} &
  \multicolumn{2}{c}{\textbf{SS}} &
  \multicolumn{2}{c}{\textbf{MG}} &
  \multicolumn{2}{c}{\textbf{WD}} &
  \multicolumn{2}{c}{\textbf{Average}} \\
  \cmidrule(lr){2-3} \cmidrule(lr){4-5} \cmidrule(lr){6-7} \cmidrule(lr){8-9} \cmidrule(lr){10-11} \cmidrule(lr){12-13}
  &
  \textbf{Acc} & \textbf{F1} &
  \textbf{Acc} & \textbf{F1} &
  \textbf{Acc} & \textbf{F1} &
  \textbf{Acc} & \textbf{F1} &
  \textbf{Acc} & \textbf{F1} &
  \textbf{Acc} & \textbf{F1} \\
  \midrule


  \multicolumn{13}{l}{\textbf{Specialist Deep Learning Models}} \\[1pt]
  CarbonNet \citep{he2024carbonnet} &
    \underline{92.7} & 76.6 &
    \underline{95.2} & \textbf{87.5} &
    85.7 & 64.4 &
    83.1 & 52.0 &
    61.8 & 40.8 &
    83.7 & 64.3 \\

  AstroCLIP  \citep{parker2024astroclip} &
    92.4 & 74.7 &
    94.0 & 84.1 &
    88.7 & 62.4 &
    80.5 & 52.3 &
    80.9 & \underline{48.8} &
    87.3 & 64.5 \\
  
  GaiaNet \citep{ye2025gaianet} &
    91.1 & 67.2 &
    \textbf{95.4} & \underline{87.1} &
    89.4 & 70.3 &
    85.5 & 51.8 &
    \textbf{83.8} & 48.2 &
    \textbf{89.0} & 64.9 \\

  \midrule

  \multicolumn{13}{l}{\textbf{Proprietary VLMs}} \\[1pt]

  GPT-4.1 \citep{openai2025gpt4-1}  &
    57.9 & 27.7 &
    51.9 & 31.7 &
    60.2 & 36.1 &
    20.6 & 29.2 &
    52.3 & 24.4 &
    48.6 & 29.8 \\
  
  o3 \citep{o3}  &
    88.8 & 57.1 &
    87.1 & 53.1 &
    88.1 & 53.3 &
    81.7 & 60.0 &
    \underline{81.9} & 37.8 &
    85.5 & 52.3 \\
  
  \midrule
  \multicolumn{13}{l}{\textbf{Open-source VLMs}} \\[1pt]

  Qwen2.5-VL-3B \citep{bai2025qwen25vl} &
    30.6 & 25.8 &
    25.3 & 20.0 &
    19.0 & 26.9 &
    17.8 & 28.7 &
    48.2 & 28.6 &
    28.2 & 26.1 \\

  Qwen2.5-VL-7B \citep{bai2025qwen25vl} &
    28.4 & 25.4 &
    59.8 & 31.5 &
    19.4 & 27.3 &
    19.7 & 29.0 &
    25.8 & 28.1 &
    30.6 & 28.3 \\

  Qwen3-VL-8B \citep{bai_qwen3-vl_2025} &
    39.7 & 33.6 &
    29.8 & 27.0 &
    29.4 & 28.4 &
    22.9 & 27.3 &
    52.9 & 19.7 &
    34.9 & 27.2 \\

  Qwen3-VL-30B-A3B \citep{bai_qwen3-vl_2025} &
    22.9 & 31.3 &
    19.8 & 29.5 &
    17.5 & 26.3 &
    35.1 & 27.4 &
    65.3 & 34.4 &
    32.1 & 29.8 \\

  S1-VL-32B \citep{s1vl2026} &
    34.6 & 29.1 &
    28.8 & 31.4 &
    25.8 & 27.5 &
    24.7 & 28.4 &
    60.6 & 23.1 &
    34.9 & 27.9 \\

  S1-VL-32B-SFT \citep{s1vl2026} &
    85.9 & 60.7 &
    54.8 & 42.8 &
    64.6 & 43.7 &
    73.2 & 36.3 &
    63.3 & 27.4 &
    68.4 & 42.2 \\

  \midrule
  \multicolumn{13}{l}{\textbf{Ours}} \\[1pt]



  \rowcolor{speco3blue}
  Spec-o3-3B &
    92.0 & \underline{80.7} &
    88.0 & 75.9 &
    \underline{92.6} & \underline{84.0} &
    \textbf{93.0} & \underline{82.7} &
    78.5 & 43.2 &
    \underline{88.8} & \underline{73.3} \\[-2pt]

  \scriptsize\hspace{12pt}\emph{$\Delta$ vs Qwen2.5-VL-3B} &
    \scriptsize\color{red} +61.4 & \scriptsize\color{red} +54.9 &
    \scriptsize\color{red} +62.7 & \scriptsize\color{red} +55.9 &
    \scriptsize\color{red} +73.6 & \scriptsize\color{red} +57.1 &
    \scriptsize\color{red} +75.2 & \scriptsize\color{red} +54.0 &
    \scriptsize\color{red} +30.3 & \scriptsize\color{red} +14.6 &
    \scriptsize\color{red} +60.6 & \scriptsize\color{red} +47.2 \\
  
  \rowcolor{speco3blue}
  Spec-o3-7B &
    \textbf{93.1} & \textbf{81.0} &
    92.5 & 80.2 &
    \textbf{94.2} & \textbf{84.5} &
    \underline{90.6} & \textbf{83.4} &
    73.2 & \textbf{53.6} &
    88.7 & \textbf{76.5} \\[-2pt]

  \scriptsize\hspace{12pt}\emph{$\Delta$ vs Qwen2.5-VL-7B} &
    \scriptsize\color{red} +64.7 & \scriptsize\color{red} +55.6 &
    \scriptsize\color{red} +32.7 & \scriptsize\color{red} +48.7 &
    \scriptsize\color{red} +74.8 & \scriptsize\color{red} +57.2 &
    \scriptsize\color{red} +70.9 & \scriptsize\color{red} +54.4 &
    \scriptsize\color{red} +47.4 & \scriptsize\color{red} +25.5 &
    \scriptsize\color{red} +58.1 & \scriptsize\color{red} +48.2 \\

  \bottomrule
  \end{tabular*}
  }
\end{table*}

\section{Experiments}
\label{sec:exp}

\subsection{SpecVI-Bench}
\label{sec:data_curation}

Most existing spectral datasets are designed for the candidate screening stage and therefore use stratified sampling to construct negative examples \citep{tan2025robust,he2024carbonnet,tan2022robust}. These negatives do not reflect the vetting stage, in which contaminants can closely resemble the true targets. To evaluate and train models under this high-confusion vetting setting, we construct SpecVI-Bench.

SpecVI-Bench comprises the same five rare-object categories as in the cold-start phase, each formulated as an independent inspection task.
For each task, positive samples are taken from the corresponding official catalogs. 
To approximate the inspection stage, we construct hard negatives using a rejection-sampling procedure. 
Specifically, for each task we train a weak classifier and then sample spectra from the full LAMOST archive, retaining only sources that are not listed in the corresponding official catalogs and whose predicted positive probability exceeds 
0.8. 
On average, the rejection-sampling acceptance rate is 1.74\%.
We finally create approximately balanced training splits and deliberately imbalanced test splits to reflect the rarity of true targets in practice. 
To avoid contamination between training demonstrations and evaluation, we ensure that spectra used for cold-start trajectory construction are excluded from the SpecVI-Bench test split.
The detailed statistics are provided in the Appendix \ref{appx:specvi_stats}.

\subsection{Experimental Setup}

\paragraph{Datasets}
We evaluate on four datasets.
SpecVI-Bench is the main benchmark.
We further test generalization on the Cross-Survey set from SDSS and DESI, which tests robustness to survey-specific instrumental differences, and the Cross-Task set with unseen target categories, which evaluates transfer to new inspection tasks.
To further probe generalization limits, we additionally construct an Extreme-Case set covering quasi-stellar objects (QSO, a more complex unseen type with blended spectral morphology) and low-SNR cataclysmic variables (severely degraded signal quality, SNR $< 5$).
Survey backgrounds are provided in Appendix~\ref{app:survey_background}, and dataset details are reported in Appendix~\ref{sec:appendix_datasets}.

\paragraph{Baselines}
We benchmark Spec-o3 against three distinct categories of baselines: (1) Specialist Deep Learning Models, where we adapted and fine-tuned CarbonNet \citep{he2024carbonnet}, GaiaNet \citep{ye2025gaianet} (task-specific architectures), and AstroCLIP \citep{parker2024astroclip} (a large-scale pre-trained spectral foundation model) on the SpecVI-Bench training split; (2) Proprietary VLMs, including GPT-4.1 \citep{openai2025gpt4-1} and o3 \citep{o3}; (3) Open-Source VLMs, covering Qwen2.5-VL \citep{bai2025qwen25vl} (base models for Spec-o3), Qwen3-VL \citep{bai_qwen3-vl_2025}, and S1-VL-32B \citep{s1vl2026}, a 32B VLM post-trained on multi-discipline scientific data including astronomical observations, for which we also report a variant with our cold-start applied. All VLM baselines share Spec-o3's visualization tool and prompt for fair comparison. Specialist DL baselines target upstream screening rather than downstream vetting and thus serve as a performance reference, with a paradigm comparison in Appendix~\ref{appx:vlm_vs_dl}.

\paragraph{Metrics}
We report Acc and positive-class F1 for each task, with macro-averages across tasks. Tests are imbalanced, so F1 is primary and Acc is complementary.
For VLMs, we require the final decision to be \texttt{\textbackslash boxed\{YES\}} or \texttt{\textbackslash boxed\{NO\}}, and we compute metrics by exact matching of this output.

\paragraph{Training Details}
We use Qwen2.5-VL-3B and Qwen2.5-VL-7B as our base models. The training phase is conducted using 8 $\times$ NVIDIA H100 GPUs. The RL stages employ GRPO framework with 8 rollouts per question, limiting the maximum number of tool calls to 8 per trajectory. Further analysis of the training dynamics and convergence behavior is provided in Appendix \ref{appx:training_dynamics}.

\begin{table*}[t]
  \centering
  \caption{\textbf{Performance on unseen surveys}. We report F1 for CV, CS, and MG because SS and WD do not have sufficient samples. The LAMOST column is the training-survey reference, averaged over the same task subset on SpecVI-Bench. Best and second-best results are in \textbf{bold} and \underline{underlined}.}
  \label{tab:cross_survey}
  \small
  {
  \setlength{\tabcolsep}{6pt}
  \begin{tabular}{l | cccc | cccc | c}
    \toprule
    \multirow{2}{*}{\textbf{Model}} &
    \multicolumn{4}{c}{\textbf{SDSS}} &
    \multicolumn{4}{c}{\textbf{DESI}} &
    \textbf{LAMOST} \\
    \cmidrule(lr){2-5} \cmidrule(lr){6-9} \cmidrule(lr){10-10}
    &
    \textbf{CV} & \textbf{CS} & \textbf{MG} & \textbf{Avg} &
    \textbf{CV} & \textbf{CS} & \textbf{MG} & \textbf{Avg} &
    \textbf{Avg} \\
    \midrule

    \multicolumn{10}{l}{\textbf{Specialist Deep Learning Models}} \\[1pt]
    CarbonNet \citep{he2024carbonnet} & 49.8 & 66.5 & 46.9 & 54.4 & 49.6 & 64.1 & 58.2 & 57.3 & 72.0 \\

    AstroCLIP \citep{parker2024astroclip} & 53.9 & 64.1 & 46.4 & 54.8 & 48.3 & 53.9 & 50.8 & 51.0 & 70.4 \\

    GaiaNet \citep{ye2025gaianet} & 58.5 & 67.1 & 41.3 & 55.6 & 48.3 & 53.8 & 43.7 & 48.6 & 68.7 \\

    \midrule
    \multicolumn{10}{l}{\textbf{Proprietary VLMs}} \\[1pt]
    GPT-4.1 \citep{openai2025gpt4-1} & 29.4 & 33.4 & 28.8 & 30.5 & 28.9 & 34.4 & 31.8 & 31.7 & 29.5 \\
    o3 \citep{o3}         & 59.8 & 54.3 & 57.1 & 57.1 & 60.2 & 51.4 & 56.8 & 56.1 & 56.7 \\

    \midrule
    \multicolumn{10}{l}{\textbf{Open-source VLMs}} \\[1pt]
    Qwen2.5-VL-3B \citep{bai2025qwen25vl}  & 22.6 & 25.3 & 25.4 & 24.4 & 23.3 & 30.8 & 23.1 & 25.7 & 24.8 \\
    Qwen2.5-VL-7B \citep{bai2025qwen25vl} & 26.7 & 25.9 & 27.2 & 26.6 & 25.9 & 33.7 & 30.6 & 30.1 & 28.6 \\
    Qwen3-VL-8B \citep{bai_qwen3-vl_2025} & 30.9 & 26.9 & 29.8 & 29.2 & 33.8 & 23.9 & 30.4 & 29.4 & 29.3 \\
    Qwen3-VL-30B-A3B \citep{bai_qwen3-vl_2025} & 32.1 & 34.4 & 29.8 & 32.1 & 31.5 & 31.1 & 30.2 & 30.9 & 29.4 \\

    \midrule
    \multicolumn{10}{l}{\textbf{Ours}} \\[1pt]
    \rowcolor{speco3blue} Spec-o3-3B & \underline{79.3} & \underline{76.4} & \underline{76.3} & \underline{77.3} & \underline{76.1} & \underline{70.1} & \underline{74.5} & \underline{73.6} & \underline{79.8} \\
    \rowcolor{speco3blue} Spec-o3-7B & \textbf{84.9} & \textbf{79.8} & \textbf{78.5} & \textbf{81.1} & \textbf{82.6} & \textbf{72.8} & \textbf{76.7} & \textbf{77.4} & \textbf{81.5} \\
    \bottomrule
  \end{tabular}
  }
\end{table*}

\subsection{Main Results}

Table \ref{tab:main_results} reports the in-distribution results on the SpecVI-Bench. 
Spec-o3 achieves state-of-the-art macro-average performance across five tasks.
Notably, Spec-o3-7B attains a macro-average F1 score of 76.5\%, substantially surpassing the proprietary o3 (52.3\%), its base model Qwen2.5-VL-7B (28.3\%), and the larger science-specialized S1-VL-32B-SFT (42.2\%). Our two-stage post-training strategy delivers pronounced performance gains, elevating the macro-average F1 score from 26.1\% to 73.3\% (+47.2\%) for the 3B model and from 28.3\% to 76.5\% (+48.2\%) for the 7B model.
These substantial improvements underscore the effectiveness of our framework in bridging the reasoning gap for spectral visual inspection.

Spec-o3 also outperforms specialist deep learning baselines fine-tuned on training split of SpecVI-Bench. While GaiaNet attains a similar average accuracy (89.0\%), its macro-average F1 (64.9\%) is notably lower than Spec-o3-7B (76.5\%), indicating a better precision-recall balance on imbalanced dataset. 
These results position Spec-o3 as a reliable autonomous agent for the spectral inspection, with the additional benefit of interpretable trajectories.

\subsection{Generalization Evaluation}

\paragraph{Cross-survey Generalization} 
We evaluate zero-shot performance on the Cross-Survey set from SDSS and DESI without any adaptation (Table \ref{tab:cross_survey}). Specialist deep learning models show strong degradation under these instrumental shifts. From the LAMOST reference average, their average F1 drops by -17.6\% to -13.1\% on SDSS and -20.1\% to -14.7\% on DESI, which suggests reliance on survey-specific artifacts. Spec-o3-7B remains stable with 81.1\% on SDSS and 77.4\% on DESI, which is close to its in-distribution reference of 81.5\% and well above o3 at 56.1\%. These results suggest that Spec-o3 relies on diagnostic spectral evidence that transfers across telescope projects.

\paragraph{Cross-task Generalization} 
We evaluate zero-shot performance on three unseen inspection tasks (O-, B-, and A-type spectra), where specialist deep learning models are inapplicable (Table \ref{tab:cross_task}). 
Despite the significant visual disparity between these new tasks and the training tasks, Spec-o3 demonstrates remarkable transferability. Spec-o3-7B achieves an average F1 of 76.4\%, which is +15.5\% over o3 and +45.9\% over Qwen2.5-VL-7B. Spec-o3-3B reaches 74.4\%, which is +13.5\% over o3 and +47.3\% over Qwen2.5-VL-3B. These results confirm that Spec-o3 has learned a generic, tool-assisted inspection paradigm that identifies diagnostic visual evidence defined in instructions, rather than merely memorizing specific task distributions.

\paragraph{Extreme-Case Generalization}
We evaluate on the Extreme-Case set to further probe generalization limits (full results in Appendix~\ref{appx:extreme_results}).
On the QSO subset, Spec-o3-7B achieves 71.9\% F1 in a zero-shot setting, far surpassing o3 (47.5\%) and the Qwen2.5-VL-7B base model (16.9\%). A brief incremental SFT with only 96 synthetic trajectories further improves the 3B model from 68.4\% to 73.6\%, nearly closing the gap to its in-distribution reference.
On the Low-SNR subset, Spec-o3-7B retains 78.2\% F1, only $\sim$3 points below its standard-SNR performance (81.0\%).
These results indicate that the learned inspection methodology transfers effectively to both unseen spectral complexity and noisy observations.

\begin{table}[t]
  \centering
  \small
  \caption{\textbf{Performance on the Cross-Task testset} with unseen types (O, B, A), with F1 reported. Deep learning baselines are omitted because they lack this cross-task transfer capability, and best/second-best are in \textbf{bold} and \underline{underlined}.}
  \label{tab:cross_task}
  \begin{tabular}{l | ccc | c}
    \toprule
    \textbf{Model} & \textbf{O} & \textbf{B} & \textbf{A} & \textbf{Average} \\
    \midrule
    GPT-4.1 & 32.2 & 34.1 & 33.0 & 33.1 \\
    o3         & 57.6 & 64.2 & 60.8 & 60.9 \\

    \midrule

    Qwen2.5-VL-3B & 26.0 & 28.1 & 27.2 & 27.1 \\
    Qwen2.5-VL-7B & 29.4 & 31.8 & 30.3 & 30.5 \\
    Qwen3-VL-8B & 30.7 & 32.2 & 30.8 & 31.2 \\
    Qwen3-VL-30B-A3B & 32.0 & 33.7 & 32.7 & 32.8 \\
    
    \midrule

    \rowcolor{speco3blue} Spec-o3-3B & \underline{73.5} & \underline{72.6} & \underline{77.1} & \underline{74.4} \\
    \rowcolor{speco3blue} Spec-o3-7B & \textbf{77.0} & \textbf{73.1} & \textbf{79.1} & \textbf{76.4} \\
    \bottomrule
  \end{tabular}
\end{table}

\begin{figure*}[t]
  \centering
  \includegraphics[width=\linewidth]{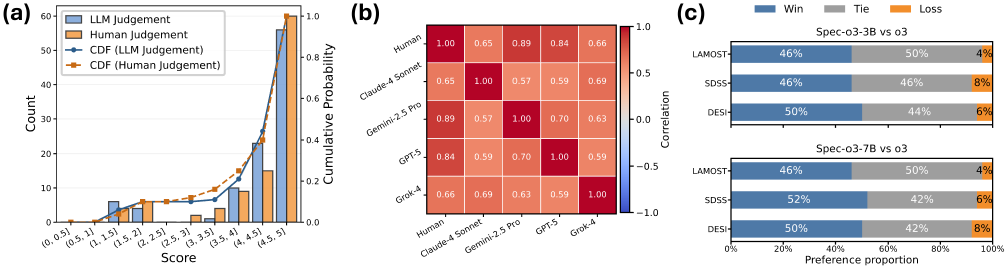}
  \caption{\textbf{(a) Quality distribution.} Score distributions and Cumulative Distribution Functions are compared between human and LLM judges to evaluate reasoning trajectories. \textbf{(b) Rating consistency.} The heatmap displays Spearman correlation coefficients between human experts and four LLM judges. \textbf{(c) Pairwise preference.} Human expert preferences on explanation quality are compared between Spec-o3 and o3 across three datasets.}
  \label{fig:human_llm_correlation}
\end{figure*}

\subsection{Reliability Judgement}
\label{sec:reliability}

To assess the reliability of Spec-o3's explanations, we evaluate the reasoning trajectories with both human experts and automated LLM Judges.
We randomly sampled 100 reasoning trajectories (50 each from Spec-o3-3B and Spec-o3-7B). Six astronomers with spectroscopy expertise rated each trajectory for coherence and physical consistency on a discrete 0-5 scale (see Appendix \ref{sec:appendix_human_eval} for comprehensive evaluation details). 
In parallel, we used four proprietary models (GPT-5, Gemini-2.5-Pro, Claude-4-Sonnet, and Grok-4) as LLM Judges to score the identical trajectories.

Figure \ref{fig:human_llm_correlation}(a) compares the score distributions and Cumulative Distribution Function (CDF) from human experts and LLM Judges, showing close agreement and a strong concentration of high scores, which suggests that Spec-o3 typically produces coherent and physically consistent explanations.
Figure \ref{fig:human_llm_correlation}(b) further reports Spearman correlations between the human scores and each LLM Judge, indicating substantial rater consistency.

Beyond absolute scores, Figure \ref{fig:human_llm_correlation}(c) presents a pairwise preference study: for each survey data, we randomly sample 50 cases and ask astronomers to compare explanation quality between Spec-o3 and o3, labeling each comparison as win/tie/loss. Across datasets, Spec-o3 is preferred at least as often as o3, with a low loss rate.
Together, these results support using LLM Judges as a scalable quality filter (e.g., for ranking or flagging low-confidence trajectories) to reduce expert workload in large surveys.

\subsection{Ablation Study}
\label{sec:ablation}

\paragraph{Impact of Two-Stage Training.} 
We conduct ablation studies to analyze the specific contributions of cold-start SFT and agentic reinforcement learning (RL), with results summarized in Table \ref{tab:ablation_main}. 
We observe that applying pure RL (\#1) yields only limited improvements, as the model lacks the foundational spectral interpretation priors necessary for effective tool utilization. 
While cold-start SFT alone (\#2) achieves performance comparable to pure RL using only sparse expert trajectories, its performance remains bounded by the scarcity of high-quality demonstrations and its weak generalization \citep{chu_sft_2025}. 
Notably, integrating both stages (\#0) triggers a substantial performance leap, nearly doubling the F$_1$ scores compared to the single-stage baselines (\#1, \#2).
This confirms that injecting domain priors via cold start is a prerequisite for RL to effectively optimize tool usage strategies and unlock robust performance.

\paragraph{Impact of Tool Access.}
To enable fine-grained inspection, we provide a visualization tool for on-demand re-rendering of local wavelength regions.
To validate the necessity of this component, we evaluated a variant (\#4) trained without tool access, forcing the model to rely solely on textual Chain-of-Thought (CoT). 
As shown in Table \ref{tab:ablation_main}, removing tool access results in a significant performance degradation for both model sizes, despite the application of full SFT and RL procedures. 
This decline shows that a static global view is insufficient for reliable verification, making interactive re-rendering of local evidence essential for detecting subtle diagnostic features.

\begin{table}[t]
  \centering
  \small
  \caption{Ablation analysis of training stages and tool availability on SpecVI-Bench. Scores denote the average F1 across five tasks.}
  \label{tab:ablation_main}
  \begin{tabular}{c|ccc|cc}
    \toprule
    \textbf{\#} & \textbf{SFT} & \textbf{RL} & \textbf{Tool} & \textbf{3B-Based} & \textbf{7B-Based} \\
    \midrule
    0 & \color{green}\ding{51} & \color{green}\ding{51} & \color{green}\ding{51} & 73.3 & 76.5 \\
    \midrule
    1 & \color{red}\ding{55} & \color{green}\ding{51} & \color{green}\ding{51} & 35.7 \scriptsize\color{red}(-37.6) & 40.5 \scriptsize\color{red}(-36.0) \\
    2 & \color{green}\ding{51} & \color{red}\ding{55} & \color{green}\ding{51} & 33.1 \scriptsize\color{red}(-40.2) & 41.6 \scriptsize\color{red}(-34.9) \\
    3 & \color{red}\ding{55} & \color{red}\ding{55} & \color{green}\ding{51} & 26.1 \scriptsize\color{red}(-47.2) & 28.3 \scriptsize\color{red}(-48.2) \\
    \midrule
    4 & \color{green}\ding{51} & \color{green}\ding{51} & \color{red}\ding{55} & 43.5 \scriptsize\color{red}(-29.8) & 55.8 \scriptsize\color{red}(-20.7) \\
    \bottomrule
  \end{tabular}
\end{table}

\paragraph{Data Efficiency of Cold Start.}
Cold start requires $\sim$1k high-quality expert trajectories, introducing a non-trivial expert-in-the-loop bottleneck for extending to new tasks or surveys.
We conduct two experiments to address this (details in Appendix~\ref{appx:data_efficiency}).
First, reducing the total SFT data from $\sim$1k to $\sim$200 trajectories across all five tasks yields only a minor performance drop (e.g., CV F1: 80.7 $\to$ 77.8 for the 3B model).
Second, we test a synthetic trajectory generation pipeline for QSO, a type not covered in training. Seeding the generator with just 3 expert examples raises the LLM-filter pass rate from 25\% to 78\%, and adding answer guidance further pushes it to 98\%.
These results show that the annotation demand can be substantially reduced, and expert trajectories can effectively bootstrap data creation for new types.

\subsection{Inference Efficiency}
\label{sec:efficiency}
Beyond accuracy, a practical vetting agent must keep pace with the data deluge that motivates this work, so we assess whether Spec-o3's iterative inference remains viable for large-scale deployment.
Although tool calling incurs per-sample overhead relative to single-pass classifiers, Spec-o3 targets high-throughput catalog construction rather than real-time single-sample processing, a regime that fully exploits modern GPU parallelism.
With efficient inference engines (e.g., vLLM \citep{vllm} and SGLang \citep{sglang}) and asynchronous tool execution, Spec-o3 processes spectra at $\sim$0.2 seconds per sample on an 8$\times$H100 server (including tool rendering), yielding a $\sim$50$\times$ throughput gain over expert manual inspection ($\sim$10 seconds per spectrum).
This efficiency makes Spec-o3 practical for vetting at the scale of next-generation spectroscopic surveys, where candidate volumes reach hundreds of thousands of spectra.

\section{Conclusion}

In this work, we introduce Spec-o3 to mitigate the manual inspection bottleneck in modern spectroscopic survey workflows.
Spec-o3 is a tool-augmented vision-language agent that performs interleaved multimodal chain-of-thought reasoning in an astronomer-like inspection loop.
Trained with a two-stage post-training strategy, Spec-o3 achieves state-of-the-art performance and remains robust under distribution shifts.
At inference time, it processes spectra at $\sim$0.2\,s per sample on an 8$\times$H100 server, delivering a $\sim$50$\times$ throughput gain over expert manual inspection.
Human evaluations confirm that its inspection trajectories are physically consistent and aligned with expert reasoning, positioning Spec-o3 as a scalable solution for the data deluge in future large-scale spectroscopic surveys.

\section{Acknowledgements}

This work is supported by the Strategic Priority Research Program of Chinese Academy of Sciences under Grant XDA0480302 and the National Key R\&D Program of China under Grant No.~2025YFF0510602.

\section{Limitations}

A limitation of this work is that our evaluation focuses on a limited set of rare-object types, and does not yet cover broader spectral subclasses or the most challenging observational conditions. In addition, we abstract expert vetting as an interactive zoom-and-reason loop with a spectral visualization tool, while real catalog construction often requires cross-matching external databases and incorporating other modalities (e.g., photometry, imaging, or time-domain evidence) for confirmation. Although outcome-based RL can scale with label-only data, our approach still relies on expert demonstration trajectories for cold start, which introduces a non-trivial barrier when extending to new tasks or surveys. Finally, we do not yet provide production-oriented risk controls such as calibration, abstention, or triage mechanisms for deferring uncertain cases to human experts.


\bibliography{custom}

\clearpage

\appendix

\section{Spectral Visualization Software}
\label{appx:software}

\begin{figure*}
  \centering
  \includegraphics[width=\textwidth]{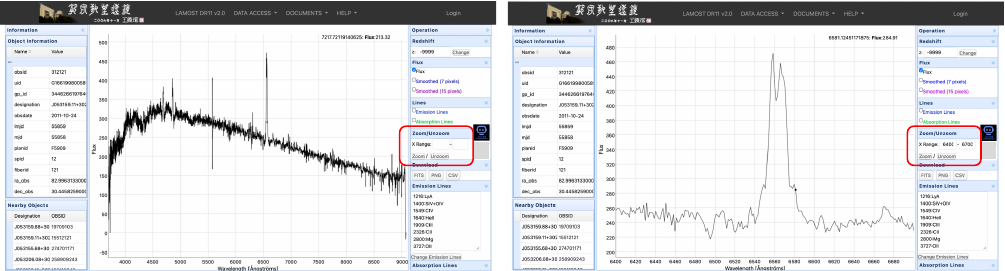}
  \caption{Screenshot of the official LAMOST spectrum viewer. The left panel lists object metadata, the center panel visualizes the spectrum, and the right panel provides interactive controls. The red box highlights the wavelength-range selection tool for zooming and re-rendering spectral details. Left: global spectrum view. Right: zoomed view re-rendered over 6400-6700\,\AA.}
  \label{fig:spectrum_tool}
\end{figure*}

Professional spectroscopic surveys typically provide dedicated visualization software to support expert-driven spectral inspection. 
Figure \ref{fig:spectrum_tool} shows a representative example from the official LAMOST spectrum viewer, which is widely used by astronomers during catalog construction and verification.

The interface is organized into three functional panels. 
The left panel displays object-level metadata, including observational identifiers, coordinates, and acquisition parameters. 
The central panel renders the spectrum as a flux-wavelength plot, enabling experts to assess the global spectral morphology and identify prominent features. 
The right panel provides interactive controls for spectral analysis, including line annotations, smoothing options, data export, and wavelength-range selection.

A key capability of this software is the ability to re-render localized wavelength regions on demand. 
By specifying a wavelength interval through the range selection tool (highlighted in red), astronomers can zoom into diagnostically relevant regions and examine fine-grained spectral structures that may not be discernible from the global view alone.
This process is typically performed iteratively: experts alternate between global inspection and multiple localized zooms to verify candidate-specific features before reaching a final decision.

The right panel of Figure \ref{fig:spectrum_tool} illustrates an example of such localized re-rendering, where the spectrum is restricted to the 6400-6700\,\AA\ range to facilitate detailed inspection of emission or absorption features in this region.
This interactive, evidence-driven workflow closely reflects real-world expert practice and motivates the tool-augmented inspection paradigm adopted in Spec-o3.

\section{Spectral Visualization Tool}
\label{appx:tool_details}

To let the VLM follow an astronomer-like, interactive inspection routine with spectral visualization software (i.e., repeatedly zooming into informative wavelength ranges), we implement a lightweight spectral visualization tool that can be invoked during inference via function calling. At the beginning of each inspection session, the tool caches the target's original one-dimensional wavelength-flux array and renders an initial spectrum plot spanning the full wavelength coverage as the starting view. In subsequent steps, the model may request a new visualization by specifying a wavelength range (e.g., $[\lambda_{\min}, \lambda_{\max}]$). The tool then slices the corresponding segment from the cached array, renders the localized spectrum view, and returns the resulting plot image to the model for the next reasoning step.

\section{SpecVI-Bench Statistics}
\label{appx:specvi_stats}

This appendix summarizes the detailed statistics and data specification of SpecVI-Bench used throughout our experiments.
SpecVI-Bench is designed to mirror the cataloging workflow in practice, where astronomers verify candidates one type at a time by inspecting spectra and deciding whether each spectrum should be accepted as the target class.
Accordingly, we formulate SpecVI-Bench as five independent binary verification tasks, one for each rare type.

Despite its binary form, SpecVI-Bench remains challenging.
Real survey spectra exhibit diverse noise patterns and instrumental artifacts, and many non-target objects can closely resemble the target types.
As a result, reliable verification typically requires substantial domain experience, even for professional astronomers.
Table \ref{tab:specvi} reports the number of positive and negative samples for each of the five rare-type spectral verification tasks, covering both training and test splits.
For each task, positive samples are drawn from the official LAMOST catalogs, while negative samples are constructed via rejection sampling to approximate the high-confusion setting encountered during expert visual inspection.
The training sets are approximately balanced to facilitate stable optimization, whereas the test sets are deliberately imbalanced, reflecting the extreme rarity of true targets in real survey scenarios.

\begin{table}[htbp]
  \centering
  \small
  \caption{Statistics of SpecVI-Bench for the five rare-type verification tasks.}
  \label{tab:specvi}
  \begin{tabular}{lcc}
  \toprule
  \textbf{Type} & \textbf{Train} (Pos / Neg) & \textbf{Test} (Pos / Neg) \\
  \midrule
  CV   & 343 / 368 & 229 / 1229 \\
  CS   & 355 / 316 & 237 / 1084 \\
  SS   & 341 / 365 & 228 / 1214 \\
  MG   & 327 / 331 & 218 / 1103 \\
  WD   & 315 / 303 & 210 / 1002 \\
  \midrule
  Total & 1681 / 1683 & 1122 / 5632 \\
  \bottomrule
  \end{tabular}
\end{table}

In addition to the binary label, each example in SpecVI-Bench provides the original 1D wavelength-flux arrays of the spectrum for on-demand visualization, together with a task prompt that specifies the verification question and the diagnostic criteria.
An example prompt is provided in Table \ref{tab:specvi_prompt_example}.

\begin{table*}[t]
\centering
\caption{Example prompt for a SpecVI-Bench sample. Here, \texttt{\textless image\textgreater} is a placeholder indicating that the model receives an input spectral image.}
\label{tab:specvi_prompt_example}
\footnotesize
\setlength{\tabcolsep}{6pt}
\renewcommand{\arraystretch}{1.08}

\setlist[itemize]{leftmargin=*, itemsep=1pt, topsep=2pt, parsep=0pt, partopsep=0pt}
\setlist[enumerate]{leftmargin=*, itemsep=1pt, topsep=2pt, parsep=0pt, partopsep=0pt}

\begin{tabular}{@{}p{0.98\textwidth}@{}}
\toprule
\textbf{Prompt Example.}\\
\midrule

\texttt{\textless image\textgreater}\\[2pt]

Analyze the provided spectrum to determine if it is a \textbf{``Cataclysmic Variables (CV)''}.\\[3pt]

\begin{itemize}
  \item \textbf{Key Indicators for CV (Check for either state):}
  \begin{itemize}
    \item \textbf{State 1: Quiescent / Low-State (Emission-Dominated)}
    \begin{itemize}
      \item \textbf{(Primary):} Strong and \emph{very broad} Hydrogen Balmer emission lines (e.g., $H\alpha$ at 6563\,\AA{}, $H\beta$ at 4861\,\AA{}). The ``broadness'' (high velocity dispersion, often $\mathrm{FWHM} > 1000~\mathrm{km\,s^{-1}}$) is the key diagnostic, indicating a high-velocity accretion disk.
      \item \textbf{(Secondary):} Broad neutral Helium (He~I) emission lines (e.g., 5876\,\AA{}, 6678\,\AA{}).
      \item \textbf{(Key Confirmation):} Presence of high-excitation \emph{ionized} Helium (He~II) emission at 4686\,\AA{}. This is a strong indicator of accretion onto a white dwarf.
      \item \textbf{(Line Profile):} Emission lines may exhibit a ``double-peaked'' profile, which is a classic signature of a rotating accretion disk viewed at high inclination.
    \end{itemize}

    \item \textbf{State 2: Outburst / High-State (Absorption-Dominated)}
    \begin{itemize}
      \item \textbf{(Primary):} Spectrum is dominated by a bright, blue continuum (looks like a hot star).
      \item \textbf{(Secondary):} The emission lines from State 1 are weak, absent, or ``filled in'', and are replaced by broad, shallow \emph{absorption} lines (primarily Balmer and He~I).
      \item \textbf{(Morphology):} The overall spectrum mimics a hot B/A-type star. The crucial difference is that the absorption lines are significantly broader, shallower, and more ``washed-out'' (or ``smeared'') than the sharp lines of a normal stellar photosphere, due to high rotational speeds and pressure broadening in the disk.
    \end{itemize}
  \end{itemize}
\end{itemize}

\vspace{2pt}
Think first, call \textbf{\texttt{spectral\_visualization\_tool}} if needed to examine features in detail, then provide your final answer. Your response must adhere to the following strict rules:
\begin{enumerate}
  \item \textbf{Overall Structure:} Your response must follow the format
  \texttt{\textless think\textgreater...\textless /think\textgreater\ \textless tool\_call\textgreater...\textless /tool\_call\textgreater\ (if a tool is used)\ \textless answer\textgreater...\textless /answer\textgreater}.
  \item \textbf{Tool Usage Constraint:} You may only make \textbf{one} tool call per turn.
  \item \textbf{Final Answer Format:} In the \texttt{\textless answer\textgreater} tag, your conclusion must be inside a \texttt{\textbackslash boxed\{\}} command (e.g., \texttt{\textbackslash boxed\{YES\}}), followed by your justification.
\end{enumerate}
\\
\bottomrule
\end{tabular}
\end{table*}

\section{Background on LAMOST, SDSS, and DESI}
\label{app:survey_background}

\paragraph{LAMOST}
The Large Sky Area Multi-Object Fiber Spectroscopic Telescope (LAMOST; also known as the Guoshoujing Telescope) is a quasi-meridian reflecting Schmidt telescope located at the Xinglong Station of the National Astronomical Observatories of China.\footnote{\scriptsize\url{https://www.lamost.org/public/instrument?locale=en}}
It provides a $5^\circ$ field of view and a highly multiplexed spectroscopic system with 4000 fibers feeding 16 spectrographs (250 fibers per spectrograph).
LAMOST operates in two survey modes: the Low-Resolution Survey (LRS) with a resolving power of $R\!\approx\!1800$ and the Medium-Resolution Survey (MRS) with $R\!\approx\!7500$.\footnote{\scriptsize\url{https://dr7.lamost.org/doc/lr-data-production-description}}

\paragraph{SDSS.}
The Sloan Digital Sky Survey (SDSS) is a long-running program that has delivered large-scale spectroscopic datasets through multiple generations of public data releases.\footnote{\scriptsize\url{https://www.sdss.org/}}
In the original SDSS spectroscopic system, observations are taken using fiber plug plates that enable 640 spectra per exposure, with a wavelength coverage of $\sim$3800--9200\,\AA\ and a resolving power of about $R\!\approx\!1800$.
Later phases employ upgraded spectrographs (e.g., the BOSS spectrographs) and continue SDSS's emphasis on broad community access to calibrated spectra and derived products.\footnote{\scriptsize\url{https://www.sdss4.org/dr17/spectro/spectro_basics/}}

\paragraph{DESI.}
The Dark Energy Spectroscopic Instrument (DESI) is a 5000-fiber multi-object spectrograph conducting the DESI Survey on the Mayall 4-meter telescope at Kitt Peak National Observatory.\footnote{\scriptsize\url{https://www.desi.lbl.gov/}}
According to the DESI instrument documentation, a wide-field corrector provides an $\sim$8 square-degree field of view, and the fibers feed 10 triple-arm spectrographs that simultaneously cover 360--980\,nm.
DESI is designed to obtain optical spectra for tens of millions of galaxies and quasars to build a 3D map of the Universe and constrain the physics of cosmic acceleration.\footnote{\scriptsize\url{https://www.desi.lbl.gov/}}

\section{Generalization Datasets Details}
\label{sec:appendix_datasets}

To evaluate the generalization ability of Spec-o3, we construct three groups of evaluation datasets: cross-survey, cross-task, and extreme-case datasets.

\paragraph{Cross-Survey Datasets.}
The cross-survey evaluation is conducted on spectra from SDSS and DESI, which differ from LAMOST in instrumentation, data reduction pipelines, and observational conditions.
Starting from the test split of SpecVI-Bench samples constructed on LAMOST, we perform cross-matching based on sky coordinates to identify corresponding observations in the SDSS and DESI archives.
A matching radius of 3 arcseconds is adopted, which is a commonly used tolerance for cross-survey astrometric matching.
For each verification task, we retain approximately 50 positive samples and 250 negative samples after matching.
Because we could not obtain enough matched samples for SS and WD, we only evaluate cross-survey generalization on CV, CS, and MG.

\paragraph{Cross-Task Datasets.}
To assess zero-shot generalization to unseen inspection tasks, we additionally construct cross-task datasets targeting O-, B-, and A-type spectra.
Positive samples for each stellar type are drawn from the corresponding official catalogs released by LAMOST.
Negative samples are collected from the same survey to preserve realistic background contamination.
For each task, we similarly retain around 50 positive samples and 250 negative samples.

\paragraph{Extreme-Case Datasets.}
\label{appx:extreme_datasets}
The Extreme-Case set comprises two subsets targeting spectral complexity and signal degradation, respectively.
The QSO subset targets high-redshift quasars, which exhibit complex spectral morphology with broad emission lines blended with host-galaxy continuum, making them a stringent test of spectral reasoning. Following the cross-task methodology above, we construct a zero-shot test set with 50 positives and 250 negatives from the official LAMOST catalog, strictly disjoint from any training or prompting data.
The Low-SNR subset targets cataclysmic variables observed under severely degraded signal quality (SNR $< 5$). We retain 250 negatives from the catalog and simulate 50 low-SNR positives by injecting Gaussian noise into validated positive spectra.

Across all settings, these datasets are used exclusively for evaluation.
No samples from SDSS, DESI, or the O/B/A/QSO tasks are included during training, ensuring a strict zero-shot evaluation regime.

\section{VLM and Specialist DL Baselines}
\label{appx:vlm_vs_dl}

Automated rare-object cataloging in astronomy is typically organized as a two-stage pipeline, in which lightweight deep learning (DL) classifiers first screen massive spectral archives for candidate targets, and domain experts subsequently vet a much smaller candidate set to ensure catalog reliability.
The specialist DL baselines used in this paper (CarbonNet, GaiaNet, and AstroCLIP) are designed for the screening stage, whereas Spec-o3 is built to automate the downstream vetting stage.
Since the two paradigms operate at different points of the workflow and under different design goals, a direct parameter-level comparison between them is not meaningful in isolation.
Table~\ref{tab:vlm_vs_dl} contrasts the two along dimensions most relevant to catalog construction, clarifying the complementary roles that motivate the comparison reported in Table~\ref{tab:main_results}.

\begin{table*}[t]
  \centering
  \small
  \renewcommand{\arraystretch}{1.2}
  \caption{Paradigm comparison between specialist DL classifiers and Spec-o3 for rare-object cataloging.}
  \label{tab:vlm_vs_dl}
  \begin{tabular*}{\textwidth}{@{\extracolsep{\fill}}lll@{}}
    \toprule
    \textbf{Aspect} & \textbf{Specialist DL} & \textbf{Spec-o3 (VLM)} \\
    \midrule
    Workflow role & Upstream candidate screening & Downstream candidate vetting \\
    Supervision & Large-scale class labels & $\sim$1k expert inspection trajectories \\
    Model size & $<$10M parameters & 3B / 7B parameters \\
    Type coverage & One model per rare type & Single model covering all five types \\
    Generalization & Retraining required per task or survey & Zero-shot transfer across tasks and surveys \\
    Interpretability & Opaque classification scores & Tool-grounded multimodal CoT reasoning \\
    \bottomrule
  \end{tabular*}
\end{table*}

In the main experiments, specialist DL models are treated as the currently deployed astronomical standard and serve as a performance reference for Spec-o3 on SpecVI-Bench. The two paradigms are complementary rather than competitive: DL classifiers remain essential for high-throughput screening over full spectral archives, while Spec-o3 automates the vetting stage that had previously relied on manual expert inspection.

\section{Training Dynamics}
\label{appx:training_dynamics}

\begin{figure*}[t]
  \centering
  \includegraphics[width=\textwidth]{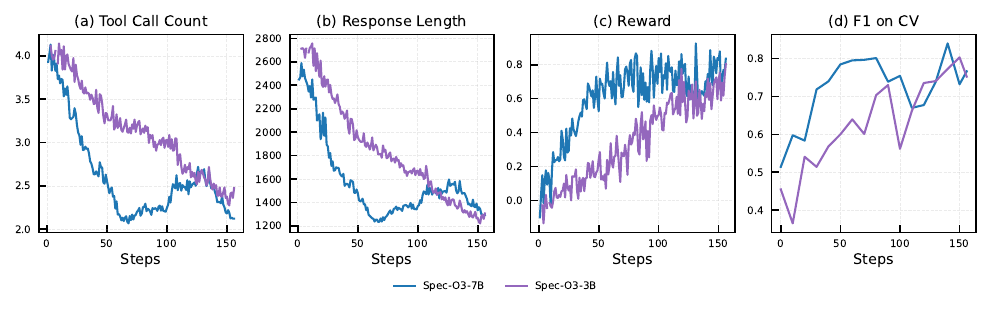}
  \caption{Training dynamics of Spec-o3-3B and Spec-o3-7B during agentic rl. Panels show (a) average number of tool calls per trajectory, (b) average response length, (c) average reward, and (d) F1 score on the CV verification set as training proceeds.}
  \label{fig:dynamics}
\end{figure*}

Figure~\ref{fig:dynamics} tracks the evolution of agent behavior during the RL stage, and Algorithm~\ref{alg:rollout} specifies the rollout procedure used for agentic RL. Reward (c) and downstream F1 (d) improve steadily for both backbones, with Spec-o3-7B converging faster and more stably than the 3B variant. The average number of tool calls per trajectory (a), in contrast, does not follow a simple monotonic trend and reveals two complementary dynamics: a policy-level exploration--exploitation transition and a process-level shift in reasoning style.

\paragraph{Exploration--exploitation transition.}
The 7B model exhibits a characteristic non-monotonic tool-call trajectory. In the initial phase (steps 0--60), it inherits the conservative, tool-intensive style of the cold-start SFT demonstrations, in which experts verify multiple spectral regions before committing to a decision; as outcome-based optimization proceeds, the policy increasingly relies on coarse evidence from the global view and the per-trajectory tool count drops. Between steps 60 and 100, easier samples have been largely solved, and the remaining cases are dominated by noisier spectra or subtle contaminants. The policy responds by increasing exploratory zooms to gather additional local evidence, producing a brief rebound in the tool-call count. Beyond step 100, the model appears to learn that further zooms bring diminishing returns on inherently ambiguous samples and converges to a more selective, stable strategy. The 3B variant exhibits a more monotonic decline, consistent with slower policy adaptation under reduced model capacity.

\paragraph{Shift in reasoning style.}
Inspection of intermediate reasoning traces reveals a concurrent behavioral change. Early-stage trajectories are generally conservative and thorough: the agent follows the task prompt, checks several diagnostic wavelength regions, and only then makes a decision---a pattern closely mirroring SFT-style imitation. Later trajectories instead exhibit an \textit{evidence-dependent stopping} pattern: the agent first examines the most diagnostic region (e.g., H$\alpha$ for CV), and if the returned view already provides strong evidence (such as a clear double-peaked emission profile), it often terminates and outputs the decision; when the first view is inconclusive, the agent continues with secondary checks. This shift from rigid guideline-following to adaptive evidence-gated stopping complements the aggregate dynamics above and further explains why the overall tool count trends downward without sacrificing accuracy.

\begin{algorithm}[t]
\caption{Rollout for agentic RL}
\label{alg:rollout}
\begin{algorithmic}[1]
\Require Spectrum array $S$ (wavelength--flux), task prompt $T_0$, policy model $\pi_\theta$, visualization tool $\mathcal{V}$, max steps $T$
\Ensure Trajectory $\tau$ for RL update

\State $\mathcal{C} \gets \textsc{CacheSpectrum}(S)$  \Comment{store the raw numeric array for this session}
\State $I_0 \gets \mathcal{V}(\mathcal{C}, \textsc{FullRange}, \textsc{label}=\varnothing)$
\State $\tau \gets [(T_0, I_0)]$

\For{$t \gets 0$ to $T-1$}
    \State $y_t \sim \pi_\theta(\tau)$ \Comment{generate next model output conditioned on history}
    \If{\textsc{IsToolCall}($y_t$)}
        \State $(\lambda_{\min}, \lambda_{\max}, \ell) \gets \textsc{ParseArgs}(y_t)$
        \State $I_{t+1} \gets \mathcal{V}(\mathcal{C}, [\lambda_{\min}, \lambda_{\max}], \textsc{label}=\ell)$
        \State $\tau \gets \tau \cup [(y_t, I_{t+1})]$ \Comment{interleave tool call and returned visualization}
    \ElsIf{\textsc{IsFinalAnswer}($y_t$)}
        \State $\tau \gets \tau \cup [(y_t)]$
        \State \textbf{break}
    \Else
        \State $\tau \gets \tau \cup [(y_t)]$ \Comment{pure text reasoning step (optional)}
    \EndIf
\EndFor

\State $r \gets \textsc{ComputeOutcomeReward}(\tau)$ \Comment{e.g., correctness and format constraints}
\State \Return $(\tau, r)$
\end{algorithmic}
\end{algorithm}

\section{Extreme-Case Generalization Details}
\label{appx:extreme_results}

We report detailed results on the two Extreme-Case subsets described in Appendix~\ref{appx:extreme_datasets}.

\paragraph{QSO Subset.}
Table~\ref{tab:qso_zero_shot} reports zero-shot results alongside the LAMOST in-distribution average as a reference. Spec-o3 shows a modest drop relative to the in-distribution reference but still substantially outperforms all baselines.

\begin{table}[h]
  \centering
  \small
  \caption{Zero-shot performance (F1) on the QSO subset.}
  \label{tab:qso_zero_shot}
  \begin{tabular}{lcc}
    \toprule
    \textbf{Model} & \textbf{QSO} & \textbf{LAMOST Avg} \\
    \midrule
    o3 & 47.5 & 52.3 \\
    Qwen2.5-VL-3B & 15.6 & 26.1 \\
    Qwen2.5-VL-7B & 16.9 & 28.3 \\
    \midrule
    \rowcolor{speco3blue} Spec-o3-3B & \underline{68.4} & \underline{73.3} \\
    \rowcolor{speco3blue} Spec-o3-7B & \textbf{71.9} & \textbf{76.5} \\
    \bottomrule
  \end{tabular}
\end{table}

To test whether the remaining gap can be recovered with minimal supervision, we perform incremental SFT on Spec-o3-3B using only 96 expert-accepted synthetic QSO trajectories generated via the few-shot pipeline described in Appendix~\ref{appx:data_efficiency}. The QSO test set remains strictly held out. As shown in Table~\ref{tab:qso_adapt}, the F1 improves from 68.4 to 73.6, nearly closing the gap to the in-distribution reference (73.3).

\begin{table}[h]
  \centering
  \small
  \caption{Lightweight adaptation on the QSO subset for Spec-o3-3B.}
  \label{tab:qso_adapt}
  \begin{tabular}{lc}
    \toprule
    \textbf{Setting} & \textbf{QSO (F1)} \\
    \midrule
    Zero-shot & 68.4 \\
    + Incremental SFT (96 synth.\ traj.) & \textbf{73.6} \\
    \bottomrule
  \end{tabular}
\end{table}

\paragraph{Low-SNR Subset.}
Table~\ref{tab:low_snr} reports results on the Low-SNR subset alongside the standard-SNR LAMOST CV reference.

\begin{table}[h]
  \centering
  \small
  \caption{Performance (F1) on the Low-SNR CV subset.}
  \label{tab:low_snr}
  \begin{tabular}{lcc}
    \toprule
    \textbf{Model} & \textbf{Low-SNR CV} & \textbf{LAMOST CV} \\
    \midrule
    o3 & 55.4 & 57.1 \\
    Qwen2.5-VL-3B & 24.1 & 25.8 \\
    Qwen2.5-VL-7B & 25.4 & 25.4 \\
    \midrule
    \rowcolor{speco3blue} Spec-o3-3B & \underline{78.0} & \underline{80.7} \\
    \rowcolor{speco3blue} Spec-o3-7B & \textbf{78.2} & \textbf{81.0} \\
    \bottomrule
  \end{tabular}
\end{table}

Under severe noise degradation, Spec-o3-7B retains an F1 of 78.2, only $\sim$3 points below its standard-SNR performance (81.0). This small drop, compared to the consistent degradation across all baselines, confirms that Spec-o3's tool-augmented inspection strategy is robust to noise, likely because its iterative zoom-and-reason loop allows verification of diagnostic features at multiple wavelength scales rather than relying on a single noisy global view.

\section{Human Evaluation Details}
\label{sec:appendix_human_eval} 

\begin{figure}[t]
  \centering
  \includegraphics[width=\linewidth]{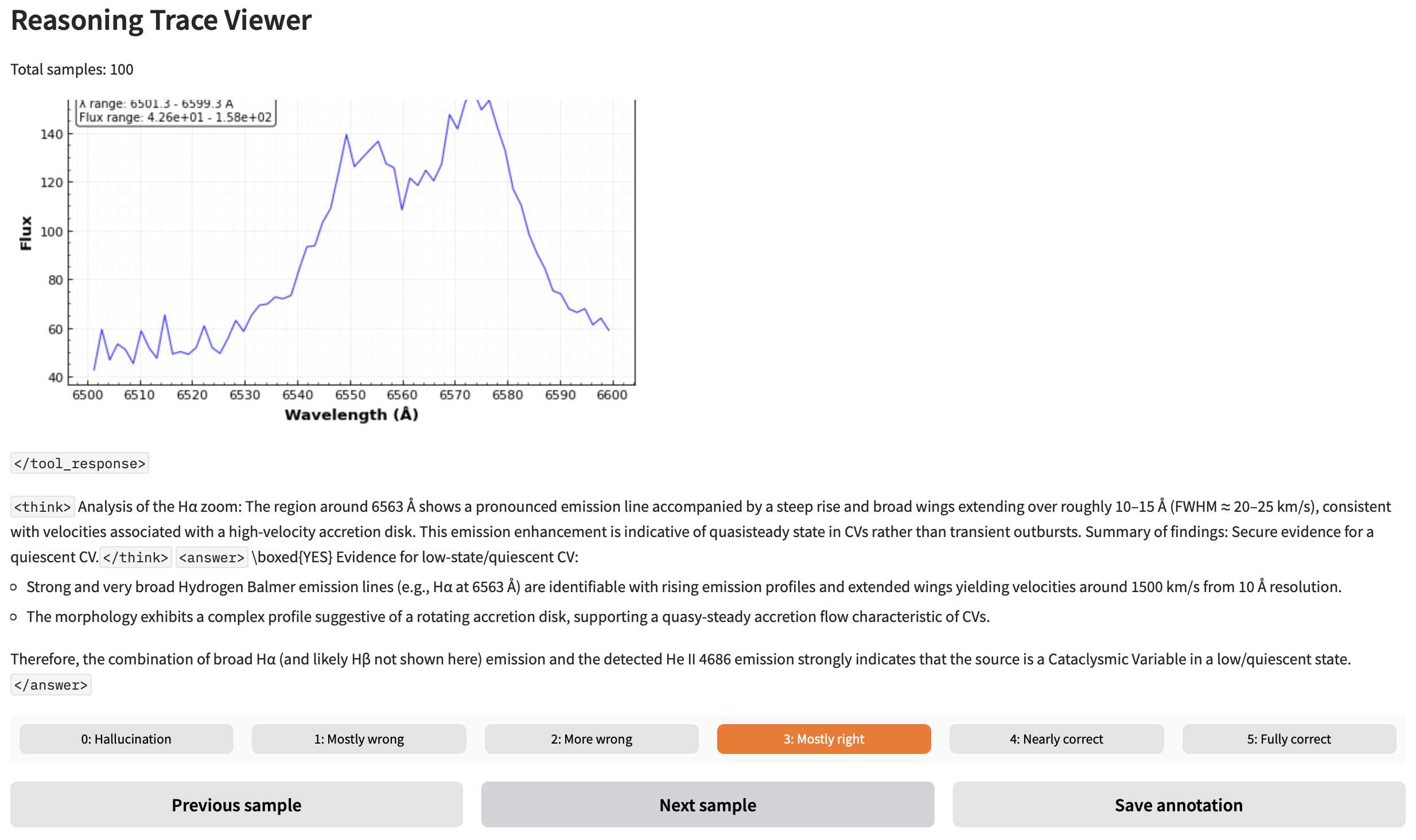}
  \caption{Custom annotation interface for expert evaluation. The interface includes a scrollable central panel that displays the full, interleaved reasoning trajectory and tool outputs. Experts review the complete chain of thought before assigning a coherence score (0-5) via the bottom control bar.}
  \label{fig:annotation_ui}
\end{figure}

\paragraph{Expert Annotators Background}
We recruited six expert astronomers for the evaluation. To ensure rigorous verification, eligibility was strictly limited to individuals holding a Ph.D. in astronomy and possessing a track record of peer-reviewed publications on spectroscopic analysis. This prerequisite guarantees that all annotators command the deep domain expertise required for accurate physical interpretation and reliable verification.

\paragraph{Scoring Rubric}
Reasoning quality was evaluated on a 0--5 scale targeting coherence and physical consistency. The criteria, summarized in Table \ref{tab:score_rubric}, span from invalid hallucinations to scientifically sound interpretations. This metric rigorously captures both the factual correctness of the answer and the internal logic of the reasoning process.

\paragraph{Annotation Interface}
To streamline the evaluation process, we developed a custom annotation interface shown in Figure \ref{fig:annotation_ui}. The interface features a scrollable reasoning viewer designed to accommodate the model's full multi-turn trajectory. This allows experts to seamlessly inspect the complete history of interleaved textual analysis and tool-generated spectral plots within a unified view. The bottom panel provides standardized controls for navigation and scoring, ensuring a consistent workflow across all annotators.

\begin{table}
  \centering
  \small
  \caption{Human evaluation rubric for reasoning coherence and physical consistency.}
  \label{tab:score_rubric}
  \begin{tabular}{>{\centering\arraybackslash}m{0.12\columnwidth} p{0.72\columnwidth}}
    \toprule
    \textbf{Score} & \textbf{Definition} \\
    \midrule
    0 &
    Complete hallucination. \newline
    The explanation is entirely incorrect or physically implausible. \\

    \midrule

    1 &
    Mostly wrong. \newline
    The majority of the explanation is incorrect or misleading. \\

    \midrule

    2 &
    More wrong than correct. \newline
    Some valid elements exist, but errors dominate the reasoning. \\

    \midrule

    3 &
    Mostly correct. \newline
    The main reasoning is sound but contains noticeable errors. \\

    \midrule

    4 &
    Nearly correct. \newline
    The explanation is largely accurate with only minor issues. \\

    \midrule

    5 &
    Fully correct. \newline
    A coherent, accurate, and physically consistent description. \\
    \bottomrule
  \end{tabular}
\end{table}

\section{Cold-Start Data Ablation Details}
\label{appx:data_efficiency}

This section provides full experimental details for the two cold-start data ablations summarized in Section~\ref{sec:ablation}.

\paragraph{Reduced SFT Data.}
We train a variant of Spec-o3-3B using only $\sim$200 SFT trajectories (reduced from $\sim$1k across all five tasks), followed by the standard RL procedure on the full training set. Table~\ref{tab:reduced_sft} reports results on two representative tasks (CV and CS). On both tasks, the reduced-data variant stays within 3 F1 points of the full-data version (CV: 77.8 vs.\ 80.7; CS: 73.2 vs.\ 75.9), while still far exceeding the base model.

\begin{table}[h]
  \centering
  \small
  \caption{Performance (F1) with reduced SFT data scale.}
  \label{tab:reduced_sft}
  \begin{tabular}{lcc}
    \toprule
    \textbf{Setting} & \textbf{CV} & \textbf{CS} \\
    \midrule
    Qwen2.5-VL-3B (Base) & 25.8 & 20.0 \\
    Spec-o3-3B (Reduced SFT, $\sim$200) & 77.8 & 73.2 \\
    Spec-o3-3B (Full SFT, $\sim$1k) & 80.7 & 75.9 \\
    \bottomrule
  \end{tabular}
\end{table}

\paragraph{Trajectory Synthesis.}
We test whether expert trajectories can bootstrap data creation for new types by collecting 100 QSO candidates (50 positive, 50 negative) and generating trajectories with GPT-5.2 under three settings: (a)~zero-shot, (b)~few-shot with 3 seed examples, and (c)~few-shot with answer guidance. Generated trajectories are filtered by Gemini-2.5-Pro and then reviewed by domain experts. As shown in Table~\ref{tab:synthetic_gen}, seed examples substantially improve the LLM-filter pass rate (from 25 to 78 out of 100), and adding answer guidance further raises it to 98. Among filtered trajectories, expert acceptance remains near-perfect across all settings, indicating that the LLM filter is well-calibrated. The 96 expert-accepted trajectories from the few-shot + answer-guided setting are used for the lightweight QSO adaptation experiment in Appendix~\ref{appx:extreme_results}.

\begin{table}[h]
  \centering
  \small
  \caption{Synthetic trajectory generation results for QSO. \textbf{LLM Filter}: pass rate out of 100 candidates. \textbf{Expert Accept}: acceptance count among LLM-filtered trajectories.}
  \label{tab:synthetic_gen}
  \setlength{\tabcolsep}{4pt}
  \begin{tabular}{@{}lcc@{}}
    \toprule
    \textbf{Method} & \textbf{LLM Filter} & \textbf{Expert Accept} \\
    \midrule
    Zero-shot & 25/100 & 24/25 \\
    Few-shot (3-shot) & 78/100 & 76/78 \\
    Few-shot + Answer-Guided & 98/100 & 96/98 \\
    \bottomrule
  \end{tabular}
\end{table}

\end{document}